\documentclass{article}

\usepackage[utf8]{inputenc}
\usepackage{url}
\usepackage{textcomp}
\usepackage{listings}
\usepackage{multicol}
\usepackage{amssymb}
\usepackage{amsmath}
\usepackage{natbib}
\usepackage{caption}
\usepackage{microtype}
\usepackage{graphicx}
\usepackage{subfigure}
\usepackage{booktabs} %
\usepackage{hyperref}

\usepackage[frozencache,cachedir=.]{minted}
\newcommand{\stf}{Swift for TensorFlow}
\newcommand{\tf}{TensorFlow}
\newcommand{\np}{NumPy}
\newcommand{\pt}{PyTorch}
\newcommand{\tensor}{\texttt{Tensor}}

\usepackage[accepted]{mlsys2021}

\mlsystitlerunning{\stf}

\begin{document}

\twocolumn[
\mlsystitle{\stf: A portable, flexible platform\\ for deep learning}

\begin{mlsysauthorlist}
\mlsysauthor{Brennan Saeta}{brain}
\mlsysauthor{Denys Shabalin}{brain}
\mlsysauthor{Marc Rasi}{brain} %
\mlsysauthor{Brad Larson}{brain}
\mlsysauthor{Xihui Wu}{brain}
\mlsysauthor{Parker Schuh}{brain}
\mlsysauthor{Michelle Casbon}{brain}
\mlsysauthor{Daniel Zheng}{brain} %
\mlsysauthor{Saleem Abdulrasool}{brain} %
\mlsysauthor{Aleksandr Efremov}{brain} %
\mlsysauthor{Dave Abrahams}{brain} %
\mlsysauthor{Chris Lattner}{formerBrain}
\mlsysauthor{Richard Wei}{formerBrain}
\end{mlsysauthorlist}

\mlsysaffiliation{brain}{Google Research, Brain}
\mlsysaffiliation{formerBrain}{Work done at Google Research, Brain}
\mlsyscorrespondingauthor{Brennan Saeta}{saeta@google.com}
\mlsyscorrespondingauthor{Denys Shabalin}{shabalin@google.com}

\mlsyskeywords{Machine Learning, Programming Languages, MLSys, Swift, Automatic Differentiation, Hardware accelerators, Edge computing}

\vskip 0.3in

\begin{abstract}
\noindent
\stf\ is a deep learning platform that scales from mobile devices to clusters of hardware accelerators in data centers.
It combines a language-integrated automatic differentiation system and multiple \tensor\ implementations within a modern ahead-of-time compiled language oriented around mutable value semantics.
The resulting platform has been validated through use in over 30 deep learning models and has been employed across data center and mobile applications.
\end{abstract}
]

\printAffiliationsAndNotice{}

\section{Introduction}
\label{section:introduction}

Deep learning has demonstrated remarkable gains in performance across diverse tasks including game playing~\cite{alphago, alphastar}, image understanding~\cite{he2016identity, inceptionv3}, natural language understanding~\cite{bert}, and beyond. %
Training modern deep learning models from scratch---a requirement during the development of new neural network architectures---requires enormous amounts of compute~\cite{openAIAndCompute}.
In practice, neural networks are trained on clusters containing up to thousands of hardware accelerators spread across a supporting data center.

Deploying machine-learned models on edge devices can deliver functionality and low latency without internet connectivity.
The popularity of modern edge devices has motivated research on models for limited hardware (compute \& memory capacity), and energy constraints~\cite{mobilenets, efficientnet}.
Additionally, many models can be fine-tuned directly on a user's device without copying personal data over a network.
While datacenter-scale training focuses on peak throughput, mobile apps optimize for startup and execution time to deliver fluid user experiences.

Day to day, machine learning practitioners are likely to develop their models on a single machine with a dedicated GPU for hardware acceleration.
Thanks to advancements in transfer learning, recent models (e.g. BERT~\cite{bert}) have been explicitly designed with pre-training in mind.
By starting from a pre-trained checkpoint, effective models can be trained on one desktop GPU.

Today, tools for training deep neural networks at datacenter scale are often embedded in dynamically typed languages~\cite{tensorflow, pytorch, jax, juliaAutodiff} and rely on Just-In-Time (JIT) compilation to obtain optimal performance.
On the other end of the spectrum, libraries for neural network inference on mobile devices are compiled before installation. 
Deploying neural networks onto devices often involves a translation step from the datacenter system to the on-device execution engine.

In this paper, we present \stf, a platform for machine learning.
The combination of Swift's Ahead-of-Time (AOT) compilation, support for mutable value semantics, and convenient syntax extended with language-integrated automatic differentation yields a surprisingly effective platform that scales from mobile phones to distributed accelerator clusters.
This work demonstrates the value of exploring beyond dynamically typed languages for deep learning. More concretely, our contributions are:

\begin{itemize}
    \item 
    \textbf{Language-integrated automatic differentiation (AD)} (Section~\ref{section:autodiff}).
    We extend Swift with a source-to-source compile-time transformation to automatically generate derivatives for arbitrary Swift functions. 
    The AD system is not coupled with the Tensor implementation; it can be used with any type that conforms to the \texttt{Differentiable} protocol and any \texttt{@differentiable} annotated function.

    \item
    \textbf{Mutable value semantics} (Section~\ref{section:value-semantics}).
    We explore mutable value semantics---a combination of value semantics with in-place mutation that supports local reasoning and referential transparency---in the context of machine learning applications.
    \stf's APIs serve as a case study of how this approach can be applied to provide simple yet powerful APIs.
\end{itemize}

We provide a broad evaluation (Section~\ref{section:evaluation}) of how the same programming model can scale across a wide range of environments from datacenter supercomputers to low-power mobile devices.

\section{Automatic Differentiation}
\label{section:autodiff}

The \stf\ project extended the Swift language to support compile-time source-to-source automatic differentiation (AD) \cite{differentiable-manifesto}.
The language extensions allow library authors to define ``differential operators,'' which are ordinary Swift higher order functions that compute derivatives of passed-in functions.
For example, we added a \texttt{gradient} function to the Swift standard library that evaluates the gradient of a scalar-valued function at a given point.
When a call to \texttt{gradient(at: 0, in: f)} is compiled, a compiler stage synthesizes a function that can be evaluated at runtime to return the derivative of \texttt{f}, which \texttt{gradient} calls.
Additionally, we support differentiation of arbitrary user-defined types so long as they satisfy a few requirements (Figure~\ref{fig:differentiable-protocol}).
Our language extensions have been contributed upstream and have been successfully merged into the main branch of the language \& compiler.

\subsection{Language extensions}

The key language building blocks for differentiation are:

\begin{itemize}
    \item The \texttt{Differentiable} protocol,\footnote{Swift protocols are similar to Haskell typeclasses.}  which encodes requirements on the parameter and return types of differentiable functions (Figure~\ref{fig:differentiable-protocol}), and
    \item The \texttt{@differentiable (A) -> B} function type family, which bundles information about how to compute the function's derivative. We call instances of these types \emph{differentiable function values}.
\end{itemize}

\begin{figure}
\begin{small}
\begin{minted}{swift}
  protocol Differentiable {
    associatedtype TangentVector
      : AdditiveArithmetic
    mutating func move(
      along direction: TangentVector)
  }
\end{minted}
\end{small}
\caption{\texttt{Differentiable} protocol definition.}
\label{fig:differentiable-protocol}
\end{figure}

Every \texttt{Differentiable} type has an associated \texttt{TangentVector} type, inspired by differential geometry.
\texttt{Differentiable} values represent points on differentiable manifolds, and \texttt{TangentVector} values represent vectors in the tangent spaces of those manifolds.
Differential geometry defines derivatives between manifolds as linear functions between the tangent spaces of these manifolds.

Consider the function \texttt{f: (A) -> Float}.
The gradient of \texttt{f} at a point is a vector in \texttt{A.TangentVector}.
When \texttt{A} is the flat manifold $\mathbb{R}^n$, \texttt{A.TangentVector} is $\mathbb{R}^n$ and we recover the gradient from multivariable calculus.

A \texttt{Differentiable} type also requires a \texttt{move} method that moves a value by the distance in the direction indicated by a \texttt{TangentVector}.
This is known as ``exponential map'' in differential geometry.

To improve ergonomics, we automatically promote functions and closures to their \texttt{@differentiable} counterparts based on their use within their surrounding module.
When type checking encounters a function value in a context that requires a differentiable function value, it inserts an implicit conversion and flags the original function for compile-time differentiation.
For example, users can pass an unannotated closure to the standard library \texttt{gradient} operator shown in Figure~\ref{fig:gradient-function}.

\begin{figure}
\begin{small}
\begin{minted}{swift}
  func gradient<A: Differentiable>(
    at x: A, 
    in f: @differentiable (A) -> Float
  ) -> A.TangentVector
\end{minted}
\end{small}
\caption{Gradient function declaration.}
\label{fig:gradient-function}
\end{figure}

\begin{figure*}
\centering
\begin{tabular}{rl}
Original function & \texttt{(A) -> B} \\
JVP (Forward mode AD) & \texttt{(A) -> (B, (A.TangentVector) -> B.TangentVector)} \\
VJP (Reverse mode AD) & \texttt{(A) -> (B, (B.TangentVector) -> A.TangentVector)} \\
\end{tabular}
\caption{The three elements of a differentiable function value and their types.}
\label{tab:bundletypes}
\end{figure*}

A differentiable function value is a bundle containing the original function value and Jacobian-vector product (JVP) and vector-Jacobian product (VJP) ``derivative function'' values (Figure~\ref{tab:bundletypes}).
Each derivative function returns a pair of the computed value together with a closure called a \textit{differential} or \textit{pullback}, respectively.
These derivative functions are taken from previous work \cite{autograd-thesis} and are particularly close to the JVP and VJP functions in \cite{jax, juliaAutodiff}.
The JVP implements forward mode differentiation, and the VJP implements reverse mode differentiation.

The AD code transformation transforms a function \texttt{f} into derivative functions that are implemented in terms of derivative functions of \texttt{f}'s callees.
The transformation recursively transforms the callees to get their derivative functions.
This recursion requires a base case of known derivative functions.
Our approach allows fully-customizable base derivative functions via a \texttt{@derivative(of:)} attribute.
Users write custom derivative functions with this attribute to register the base case, and the code transformation terminates the recursion whenever it encounters a function with a user specified custom derivative.

\subsection{Code transformation}

The differentiation code transformation operates on the Swift Intermediate Language (SIL), an intermediate representation (IR) in static single assignment form.
SIL is designed specifically for Swift's compiler transformations, which helps us to handle diverse Swift language constructs (e.g. various control flow statements, value data types, and higher-order functions).
The differentiation transformation involves the following steps:
\begin{itemize}
    \item \textbf{Activity analysis} \cite{tapenade} determines instructions of the original function that are both \textit{varied} (depend on the inputs) and \textit{useful} (contribute to the output). Such instructions are \textit{active} and need a derivative. %
    \item \textbf{Differentiability checking} detects non-differentiable instructions and emits errors and warnings (e.g. a differentiable function whose return value does not depend on differentiable arguments) that help users catch errors before execution.
    \item \textbf{Derivative synthesis} creates the derivative functions, applies AD rules to active SIL instructions, and builds the corresponding derivative SIL instructions. This step also generates code that captures callee derivatives and the control flow path (see below). 
\end{itemize}

Since AD is a compiler pass operating on the intermediate representation, the code generated by the AD system is fully amenable to the same set of compile-time optimizations as regular Swift code.

The code transformation produces the JVP and VJP discussed earlier in Figure~\ref{tab:bundletypes}.
In order to be valid for every possible input $x$, the JVP and VJP must handle all the possible control flow paths through the original function.
Our approach is based on statically-typed records corresponding to the basic blocks of the control flow graph that store intermediate state used in derivative calculations.
These records form a nested data structure of control flow branches between basic blocks that have been taken during the execution of the function.

\subsection{Contributions, limitations, \& future work}

To our knowledge, this combination of (1) requesting differentiation within the language, (2) supporting user-defined types with arbitrary tangent vector types, and (3) performing the AD code transformation at AOT-compile time was first explored in \stf.

Enzyme~\cite{enzyme} is a successor system that shares a similar set of the advantages, only targeting LLVM IR rather than SIL.
\cite{tapenade} performs the code transformation at compile time in an AOT language, but requires that the user request differentiation outside the language.
Clad~\cite{vassilev2020automatic} is a similar successor system operating on C/C++. 
Other AD systems that allow the user to request differentiation in the language trace the computation at runtime and differentiate the trace~\cite{tensorflow, autograd, jax, pytorch} or use dynamic language features to transform the code at runtime~\cite{juliaAutodiff}.

Our approach currently does not support certain capabilities that are supported in similar systems for dynamic languages, such as higher-order differentiation.
The \texttt{@differentiable} function type family would need to be extended to encode that an $n$-times differentiable function is $(n-1)$-times differentiable.
Additionally, the code transformation currently cannot transform its own output because the output makes heavy use of closure captures, which are not yet well supported.
A possible approach for differentiation of closures with captured variables is documented in \cite{diffcurry}.
Finally, there are additional language features to support, such as \texttt{enum} types.

\section{Tensors \& Lazy Tensors}
\label{section:tensor}

Deep learning models are often trained using a multi-dimensional array abstraction, colloquially called a \tensor~\cite{tensorflow, pytorch}.
A \tensor\ type includes a suite of linear algebra operations such as matrix multiplication, convolution, and element-wise arithmetic.
Because the AD system has been generalized to arbitrary types, there is no coupling between the AD system and a specific \tensor\ type.
As a result, we have found that implementing multiple different \tensor\ types allows the platform to more effectively scale to different applications.

\subsection{Na\"ive Tensor}

One of our multiple \tensor\ implementations is a single-threaded type backed by Swift arrays.
Although this type lacks hardware accelerator support, this implementation has advantages when working with small \tensor s, including portability, low computation and memory overheads, and small binary size (since the na\"ive implementation has no external dependencies).

\subsection{Eager Tensor}

Some libraries (e.g. \np) eagerly dispatch the operations of the user's program to pre-compiled functionality (termed ``kernels'') that can run on CPUs or hardware accelerators such as GPUs.
``Eager mode'' (also known as define-by-run~\cite{chainer}) is popular for research because it supports the full flexibility of the host programming language, and is known to make debugging easier.
In order to make efficient use of hardware accelerators, the kernels are dispatched to the accelerator to execute asynchronously and control is returned to the user's program before the kernel finishes.~\cite{pytorch}
As long as the user's program does not observe the contents of a \tensor, the user's program runs ahead and fills a pipeline of accelerator kernel invocations making efficient use of hardware resources.

\stf\ delivers a define-by-run machine learning environment and exposes an eager execution model to the end user. 
We provide a hardware-accelerated implementation of \tensor\ operations backed by TensorFlow's eager runtime \cite{tensorflow-eager}.
Eager \tensor\ performs asynchronous op-by-op execution that dispatches to hardware-optimized kernels under the hood.

The downside of the eager execution model is that it does not expose opportunities for operation fusion to the underlying runtime. 
This is the direct side-effect of op-by-op execution that is driven by the host programming language.

\subsection{Lazy Tensor}

Domain-specific optimizing compilers~\cite{tvm, xla, glow} have been developed to both (a) substantially improve performance of deep learning models, and (b) target domain-specific hardware accelerators such as TPUs~\cite{cloudtpu}.

Unlike eager execution, such compilers can take complete models as programs in their own domain-specific IR and generate optimized hardware-specific machine code. 
The ability to observe the complete program provides a wide horizon for optimizations such as operation-fusion.

We've developed \emph{LazyTensor} as a second implementation of our eager \tensor\ API that relies on tracing \cite{jaxtracing} to target just-in-time (JIT) compilation via HLO, the XLA compiler's \cite{xla} IR.
End-users can switch between the two implementations by specifying a device for the computation to run on: either an eager or a lazy-tracing one.

\begin{figure}
\includegraphics[width=0.479\textwidth, angle=0]{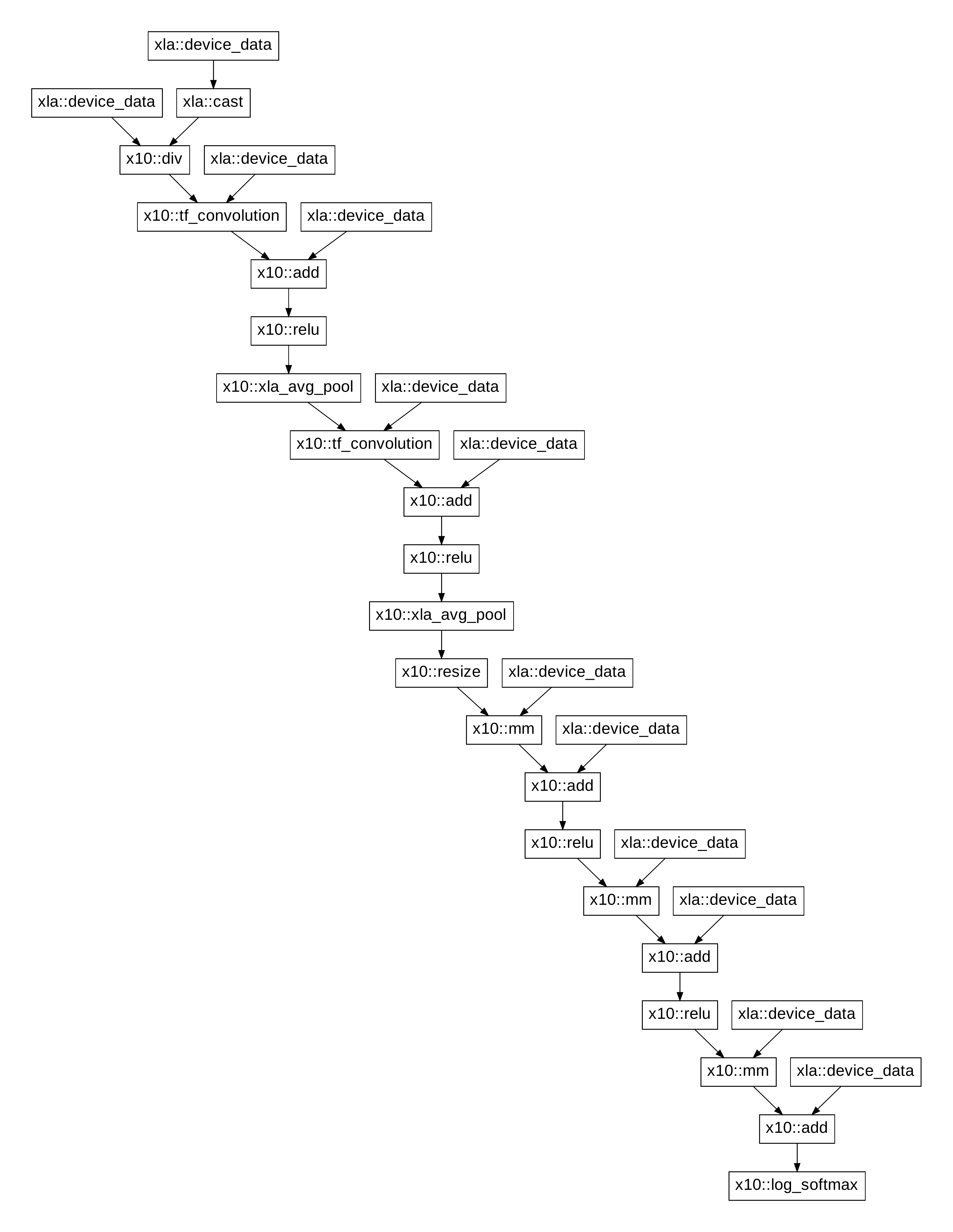}
\caption{LazyTensor trace of the LeNet-5 model's forward pass.}
\label{fig:lazytensor-trace}
\end{figure}

The biggest challenge behind the alternative implementation was not to break the illusion of eager execution.
As long as the user's program does not observe a \tensor's contents, the code cannot distinguish when a \tensor\ operation is actually executed (ignoring timing and similar side-channels).
Instead of dispatching to a fixed-set of pre-compiled kernels, the lazy \tensor\ type simply records a dynamic \textit{trace} of operations to be executed at a later time. 
Traces are represented in-memory as directed acyclic graphs (Figure~\ref{fig:lazytensor-trace}) and are transformed into an intermediate representation to perform domain-specific optimization and code generation. 

By dynamically discovering traces implicit within the user's program, \stf\ enables automatic composition of \tensor\ and non-\tensor\ computations. 
Imagine a hypothetical robotics motion planning algorithm that is composed of a neural network, a black-box CPU solver, and a second neural network.
If this motion planning algorithm is used as a subroutine for a larger \tensor-based computation, known competing approaches require the user to factor the algorithm into at least three separate subroutines and annotate a subset of the subroutines for compilation with a domain specific compiler.
With our approach, the longest possible traces are discovered dynamically and function composition is never impinged.

Through LazyTensor, \stf\ delivers the usability of eager mode \emph{and} the performance of domain-specific optimizations and code generation.
This technique is detailed in a concurrent paper with collaborators from \pt~\cite{lazyTensor}. 

\subsection{Limitations}

One of the drawbacks of the LazyTensor approach is the overhead due to re-tracing the user's program repeatedly.
Because invoking the XLA JIT is computationally expensive, trace fragments are hashed to become keys in an XLA-program cache; each unique trace is only compiled by XLA once.
Even though we reuse previously compiled traces, we still incur tracing overhead on each iteration since we support the full imperative programming model, and traces can change at any point in program execution.

Since XLA was designed with an expectation of statically known tensor shapes, minor changes in program execution such as changes in the dimensions of the input tensors can trigger recompilation. 
As a consequence, lazy tracing works best when the computation is done repeatedly over the same constant tensor dimensions on each iteration.

While constructing the trace, we fully unroll any control-flow.
This is often desirable as it allows for optimizations across loop iterations.
As a downside, this can cause the creation of large traces, which can incur significant one-time JIT compilation costs.
We expose an additional library function \texttt{LazyTensorBarrier()} that allows users to explicitly cut traces.
Fortunately, because the most common loop that is accidentally unrolled is the main training loop of a neural network, a training-loop library can automatically call \texttt{LazyTensorBarrier()} after the optimizer update step on behalf of the user.

It is left to future work to investigate automatically detecting a sufficiently large trace fragment to compile and dispatch automatically, completely relieving the user of the need for any annotations.

\subsection{Alternatives}

Prior to lazy tracing, the \stf\ project explored an approach called \emph{graph program extraction}.
The key idea behind it was to slice the user's program into two concurrent programs, one that would execute on an accelerator and one on the CPU in the user's host language, with communication channels between them for coordination.
The accelerator programs could thus be compiled into fused kernels completely ahead-of-time.

We found this approach to be quite limiting since models often rely on dynamically configured values that are only available at runtime.
For example, one may implement a complete ResNet family of models by assembling key building blocks in a configuration determined by a dynamic model variant.
A fully static implementation is limited to fusion within the individual building blocks since the final composition is not known ahead-of-time.
Alternatively, it could also attempt to pre-compile a large number of variants that exercise all possible combinations of the configuration space (which can be exponential). 

Julia's approach to XLA compilation \cite{juliaTPU} makes a different set of trade-offs.
Rather than being embedded in an AOT compiled language, they use XLA as an alternative backend to Julia's JIT compiler. 
The language provides a set of powerful runtime code generation and specialization facilities that allow it to address XLA's limited expressiveness. 

\section{Mutable Value Semantics}
\label{section:value-semantics}

\begin{figure*}
\begin{multicols}{3}
\begin{minted}{python}
# Python integer
x = 3
y = x
x += 1
print(x) # 4
print(y) # 3
\end{minted}

\columnbreak

\begin{minted}{python}
# Python list
x = [3]
y = x
x[0] += 1
print(x)  # [4]
print(y)  # [4] <===
\end{minted}

\columnbreak

\begin{minted}{swift}
// Swift array
var x = [3]
var y = x
x[0] += 1
print(x)  // [4]
print(y)  // [3]
\end{minted}
\end{multicols}
\caption{Examples of value and reference semantics in Python and Swift. In
  Python, integers have value semantics (first column), but lists and
  user-defined types have reference
  semantics: in the middle column, a mutation through \texttt{x} changes the
  value observed through \texttt{y}. In Swift, mutable value semantics are pervasive (third column).}
\label{fig:valex}
\end{figure*}

When a type has value semantics, distinct variables of that type access logically disjoint data, so mutations through one variable are observable only through that variable.\footnote{Immutable data has value semantics trivially.} This behavior should be familiar to most programmers: in almost every major language, mutation through an integer variable affects only that variable (Figure~\ref{fig:valex} column 1).

The alternative is \emph{reference semantics}, where mutation through one variable can be observed through another.  Most languages that support mutation allow only reference semantics for user-defined and aggregate types (such as arrays). However, reference semantics has well-known downsides: loss of local reasoning and referential transparency; ``spooky action at a distance''\footnote{With apologies to Einstein.} where data changes unexpectedly due to implicitly shared state (Figure~\ref{fig:valex} column 2), defensive copying, the inability to optimize code where state is not actually shared, and data races in concurrent code.  It is particularly difficult to implement mathematical transformations such as AD in systems with reference semantics.

These hazards often drive designers to adopt \emph{immutability}, producing a pure functional programming model. For example, the \tensor\ types in \tf, JAX, and XLA are immutable.  However, strict immutability can be inefficient, difficult to map onto some algorithms~\cite{sieve}, and often clashes with the programmer's mental model.\footnote{Both XLA and \tf\ have compromised their pure programming models for expressivity and performance reasons.} A common intermediate position keeps data immutable but implements state in the programming model by rebinding variables to newly-constructed immutable values.  In either case, the performance implications of making a small update\footnote{Even in pure functional programming, it's still common to create updated versions of large values. Persistent data structures~\cite{Driscoll89makingdata} address the problem in part, but cannot match the performance of contiguous arrays.} to a large value are unfavorable.

A type has \emph{mutable value semantics} when two variables of the type always have logically-disjoint values (value semantics) \emph{combined with} support for efficient part-wise mutation.
C++, Swift, and Rust are examples of popular languages that support defining such types.

In Swift, built-in types like integers, arrays, and strings have mutable value semantics, and user-defined types composed from such types automatically acquire mutable value semantics of their own.\footnote{Unless explicitly requested by using the \texttt{class} keyword.}  
Moreover, mutation always occurs via a ``unique borrow'' of the value being mutated, marked in a function signature with \texttt{inout}.  Although an \texttt{inout} parameter superficially resembles a pass-by-reference parameter, it does not introduce reference semantics because the borrow is guaranteed to be unique.  In fact, a program using \texttt{inout} can be trivially rewritten into a semantically-equivalent program using only pass-by-value (Figure~\ref{fig:inout}). Because \stf\ \tensor s have value semantics, we can perform sophisticated static analysis such as static shape tracking~\cite{tfp} even in the presence of mutation.

Swift occupies a unique position in the design space for mutation. In languages with reference semantics, users must copy defensively to avoid ``spooky action at a distance.'' Too many copies results in performance problems; too few copies results in bugs. Other languages with value semantics make copies eagerly, causing users to pass parameters by reference, thus bypassing value semantic guarantees.  In Swift, large values are copied lazily, upon mutation, and only when shared, so pass-by-value is efficient and broadly used.  The reference counting mechanism underlying this behavior is built in, enabling optimizations not possible in other languages.

Next, we show how mutable value semantics in Swift enable elegant solutions to important challenges for ML systems.

\subsection{Tensors, Layers, \& Models}

\stf\ APIs use mutable value semantics pervasively (e.g., \tensor s, models, and datasets are all mutable value types).
Unlike TensorFlow, Keras~\cite{tensorflow-keras} and \pt~\cite{pytorch}, \stf\ does not need a \texttt{Variable} type; composition of mutable value semantics and language-integrated AD allows us to use the types directly without any additional wrappers.

As an example, we provide a full definition of a variant of the LeNet-5 convolutional neural network model~\cite{LeNet} in Figure~\ref{fig:lenet}.
The LeNet model is simply a \texttt{struct} that conforms to the \texttt{Layer} protocol that composes a number of standard  layers. %
Each conforming \texttt{Layer} must provide an implementation of \texttt{callAsFunction} that defines how to apply a transformation to a given input. 
This function must be annotated as \texttt{@differentiable}, which instructs the compiler to enable automatic differentiation (discussed in Section~\ref{section:autodiff}).

\begin{figure*}
\begin{small}
\begin{minted}{swift}
  public struct LeNet: Layer {
    public var conv1 = Conv2D<Float>(filterShape: (5, 5, 1, 6), 
                                     padding: .same, activation: relu)
    public var pool1 = AvgPool2D<Float>(poolSize: (2, 2), strides: (2, 2))
    public var conv2 = Conv2D<Float>(filterShape: (5, 5, 6, 16), activation: relu)
    public var pool2 = AvgPool2D<Float>(poolSize: (2, 2), strides: (2, 2))
    public var flatten = Flatten<Float>()
    public var fc1 = Dense<Float>(inputSize: 400, outputSize: 120, activation: relu)
    public var fc2 = Dense<Float>(inputSize: 120, outputSize: 84, activation: relu)
    public var fc3 = Dense<Float>(inputSize: 84, outputSize: 10)

    public init() {}

    @differentiable
    public func callAsFunction(_ input: Tensor<Float>) -> Tensor<Float> {
        let convolved = input.sequenced(through: conv1, pool1, conv2, pool2)
        return convolved.sequenced(through: flatten, fc1, fc2, fc3)
    }
  }
\end{minted}
\end{small}
\caption{LeNet model definition.}
\label{fig:lenet}
\end{figure*} %

\subsection{Avoiding model copies}

Training a neural network model can be thought of as repeated application of a function from network parameters and a training minibatch to updated network parameters.
In code, this could be written as a function with type:
\begin{verbatim}
    (Model, Minibatch) -> Model
\end{verbatim}

Some neural network models (e.g. large transformer-based natural language models) consume a significant fraction of accelerator memory capacity for the model's weights.
When training such networks, materializing two copies of the network's parameters in memory will exceed accelerator capacity.
XLA's pure-functional, immutability-oriented core has been extended with the concept of input-output buffer aliasing to accommodate these workloads.
Unfortunately, this compromise leaks into the user programming model.
Swift's \texttt{inout} keyword (see Figure~\ref{fig:inout}) is a solution that allows us to rewrite the training function's type to be:
\begin{verbatim}
    (inout Model, Minibatch) -> Void
\end{verbatim}
The semantics are identical to the pure-functional approach, while simultaneously avoiding exceeding hardware limits so long as the \tensor\ type used by the model properly implements mutable value semantics.

As an example, an explicit training loop for the LeNet model can be written in a few lines of code shown in Figure~\ref{fig:train}.
After instantiating the model, the user takes the gradient of a differentiable function computing a loss, with respect to the model.
The returned gradients are of type \texttt{Model.TangentVector}, and can be stored or manipulated as a first-class value.
An optimizer borrows the model uniquely, and updates it in-place based on the computed \texttt{gradients}.
\begin{figure}
\begin{small}
\begin{minted}{swift}
var model = LeNet()
for epoch in epochs {
  for batch in epoch {
    let gradients = gradient(at: model) {
      model -> Tensor<Float> in
        let loss = softmaxCrossEntropy(
          logits: model(batch.data), 
          labels: batch.label)
        return loss
    }
    optimizer.update(&model, 
                     along: gradients)
  }
}
\end{minted}
\end{small}
\caption{Training LeNet on a simple dataset.}
\label{fig:train}
\end{figure}
Because both the model and its gradient are first class values, both can be manipulated (e.g. storing previous values, or computing properties of them).

\subsection{Array indexing differentiation}

The efficient gradient design goal states that computing the derivative of a function should be comparable in cost to computing the function itself.
AD systems compute derivatives of a composite function by combining the derivatives of constituent operations using the chain rule.
While pure-functional languages or intermediate representation (IRs) have formed the basis of numerous powerful \tensor-based AD systems, the derivative of the array indexing operation violates the AD efficiency goal.
The types\footnote{Without loss of generality, we assume \texttt{T == T.TangentVector} and \texttt{[T] == [T].TangentVector}. Additionally, we use a regular Swift array instead of the multi-dimensional \tensor\ type.} of the array indexing operation are:

\begin{verbatim}
  // Operation type:
  @differentiable(wrt: a)
  (a: [T], index: Int) -> T

  // Pullback type:
  (T) -> [T]
\end{verbatim}

Unfortunately, this functional AD formulation necessitates the construction of an array of all zeros\footnote{Technically, the additive identity.} except for a single non-zero entry at position \texttt{index}.
While a secondary, sparse representation for the array-type could improve performance, runtime overheads remain in practice.
Although Dex~\cite{dex} has developed an alternative approach leveraging monads and an algebraic effects system, they add complexity to the language.
A mutable-value-semantic AD formulation could instead write this derivative with the following types:

\begin{verbatim}
  // Pullback type:
  (T, inout [T]) -> Void
\end{verbatim}

This new formulation avoids materializing zeros unnecessarily, and composes correctly in the presence of additional operations.
Please see Appendix~\ref{app:arr-ad} for the complete implementation of the operations.
Further, the technique developed here applies not just to array or \tensor-operations, but additionally to derivatives of arbitrary ``big-to-small'' operations on many data types, such as manipulating a vertex within a tree or graph, or computing the partial derivative with respect to a field within an aggregate data type.

\subsection{Open questions}

Our work has demonstrated that languages oriented around mutable value semantics enable efficient solutions to important challenges for ML and \tensor-based systems.
However, work remains on how to implement a domain-specific compiler around a mutable value semantics core.
Additionally, further work is required to augment the AD system with support for \texttt{inout}-formulated derivatives, and automatically determine when to use a functional-formulation of the pullback versus the \texttt{inout} formulation.

\section{Applications and Evaluation}

\label{section:evaluation}

\stf\ brings together differentiable programming in a compiled language with high-level abstractions over heterogenous compute resources, and enables applications in several domains, including machine learning and particularly the field of deep learning.
Beyond machine learning, \stf\ has been applied to differentiable physics simulations and highly-parallel numerical calculations.

A shared repository of examples, building blocks, and benchmarks is maintained at \url{https://github.com/tensorflow/swift-models}.
As of this writing, over 30 examples exist at that location, spanning such areas as image classification, generative models, recommendation systems, and reinforcement learning.

Two recent works used \stf\ to assist in reinforcement learning research.
A simulated environment for never-ending learning called Jelly Bean World~\cite{Platanios2020Jelly} was constructed using \stf\ in order to run efficiently on low-cost hardware.
DeepMind released a framework for researching reinforcement learning in games called OpenSpiel~\cite{LanctotEtAl2019OpenSpiel} with an implementation in \stf.

\subsection{Training performance}

To evaluate the performance of \stf, we present three scenarios that demonstrate the range of hardware supported by the platform.
This includes training a deep learning model on accelerators such as Tensor Processing Units (TPUs)~\cite{tpu, cloudtpu} and Graphics Processing Units (GPUs), as well as refining simpler machine learning models on mobile devices.

\subsubsection{Training on TPUs}

TPUs are a domain-specific hardware accelerator optimized for deep learning models.
Because the XLA compiler is the only supported mechanism to generate TPU programs, \stf\ uses the LazyTensor approach (detailed above) to JIT-compile traces of \tensor\ operations.

The performance of \stf\ on TPUs was measured by training the ResNet-50 image classification network~\cite{he2016identity} on the ImageNet 2012 dataset~\cite{imagenet_cvpr09} using TPUv3-16, TPUv3-32 and TPUv3-128 clusters, shown in Table \ref{tab:s4tftpuresnetperf}.
The model was trained for 90 epochs, and both the time required as well as the post-warmup throughput in examples per second were recorded.
For comparison, a ResNet-50 model implemented in TensorFlow and in JAX  using the Flax framework~\cite{flax} were trained on TPUv3-32 under the same conditions, shown in Table \ref{tab:tpuresnetperf}.
The model, dataset, and training conditions were selected to conform to the MLPerf Training Benchmark~\cite{MLPerf}.

\begin{table*}[h]
    \centering
    \begin{tabular}{|l|r|r|r|r|}
        \hline
        \# Cores & Validation Accuracy & Training Time & Throughput & Per-Accelerator Throughput \\
        & (top-1) & (90 epochs) & (examples / s) & (examples / s / TPU core) \\
        \hline
        16 & 78.1\% & 189 minutes & 10164 & 635.25 \\
        32 & 77.7\% & 96 minutes & 20015  & 625.47 \\
        128  & 77.8\% & 25 minutes & 77726  & 607.23 \\
        \hline
    \end{tabular}
    \caption{
        \stf\ training performance for ResNet-50 on ImageNet on TPUv3 clusters.
        Per-accelerator throughput is largely maintained while scaling from a single host to 8 hosts synchronously training a single model in data-parallel fashion, demonstrating the scalability of the LazyTensor approach employed by \stf\ to target TPUs.
    }
    \label{tab:s4tftpuresnetperf}
\end{table*}

\begin{table*}[h]
    \centering
    \begin{tabular}{|l|r|r|r|}
        \hline
        Framework & Validation Accuracy & Training Time & Throughput \\
        & (top-1) & (90 epochs) & (examples / s) \\
        \hline
        JAX + Flax & 76.8\% & 90 minutes & 21258 \\
        \tf & 77.9\% & 59 minutes & 33118 \\
        \stf  & 77.7\% & 96 minutes & 20015 \\
        \hline
    \end{tabular}
    \caption{Training performance for ResNet-50 on ImageNet on a TPUv3-32 cluster. Although each system can notionally produce identical XLA HLO and thus achieve equivalent performance, some codebases have been better optimized for benchmark purposes. We include this table for completeness.}
    \label{tab:tpuresnetperf}
\end{table*}

Although \stf\ trains in approximately equivalent time as both JAX and \tf, \stf\ is the only framework with an eager programming model on TPUs. 
By contrast, both JAX and \tf\ stage the user's code in their own intermediate representations (jaxpr and GraphDef, respectively).
Note that because the \stf\ implementation achieves higher accuracy typically associated with longer training times (due to algorithmic tweaks inspired by fastai~\cite{fastaibook}), the other frameworks are at a slight advantage.

\subsubsection{Training on GPUs}

GPUs are also commonly used, so the training performance of \stf\ was evaluated on a commodity NVIDIA GTX 1080, shown in Table \ref{tab:gpuresnetperf}.
The ResNet-56 image classification network~\cite{he2016identity} was trained against the CIFAR-10 dataset~\cite{cifar10} for 10 epochs, measuring training throughput after a warmup period.
Training was performed on the TensorFlow and \pt\ frameworks using the same model, dataset, and hardware. 

\begin{table*}[h]
    \centering
    \begin{tabular}{|l|r|}
        \hline
        Framework & Throughput \\
        & (examples / s) \\
        \hline
        \pt & 2462 \\
        \tf & 2390 \\
        \stf\ (Eager Mode) & 730\\
        \stf\ (LazyTensor) & 1827\\
        \hline
    \end{tabular}
    \caption{Training performance for ResNet-56 on CIFAR-10 on an Nvidia GTX 1080 GPU.}
    \label{tab:gpuresnetperf}
\end{table*}

\stf's LazyTensor tracing significantly outperforms the operation-by-operation dispatching of its eager mode, due to the overhead inherent in the latter and the XLA compiler's ability to selectively fuse operations. \pt\ and TensorFlow both outperform \stf's LazyTensor backend on GPUs at present in part due to LazyTensor tracing overhead. Improvements in caching of LazyTensor traces should lead to increased throughput.

\subsubsection{Training on mobile devices}

Learning parameters through iterated optimization has applications beyond deep learning, such as learning knots in a polynomial spline.
Splines require orders of magnitude less computation and are thus attractive in resource constrained environments such as mobile phones.
Optimization algorithms such as backtracking line search use derivatives to determine the step direction.

Swift's AD capabilities are not tied to any underlying accelerator interface or platform, and code can be compiled to target resource-constrained hardware, such as mobile or embedded devices.
Swift is the most popular language to develop applications for Apple, Inc.'s iOS platform, and the compiler can target many other mobile or embedded hardware configurations.

As an example of this, a proprietary personalization model using splines optimized via backtracking line search was implemented in \stf.
The hardware-accelerated Swift \tensor\ type used to train on GPUs or TPUs was replaced with a CPU-only implementation backed by Swift arrays.
This model was cross-compiled to binaries intended for ARM32 and ARM64 Android devices.

A global spline model was trained on anonymized, aggregated data, and fine-tuned on a Google Pixel 3 phone using only local data.
We compare implementations of this algorithm using three distinct frameworks: TensorFlow Mobile~\cite{tensorflow}, TensorFlow Lite, and \stf.
In the case of TensorFlow Lite, two versions were tested: one with the standard available operations and another with a custom fused operation designed specifically for this model.
The spline model amplifies even slight numerical differences in calculations performed during training, so the results of all three frameworks were verified to produce control point values that matched within 1.5\% of each other.

\begin{table*}
    \centering
    \begin{tabular}{|l|r|r|r|}
        \hline
        & Training Time & Memory Usage & Binary Size \\
        Platform & (on device) & (on device) & (uncompressed) \\
        \hline
        TensorFlow Mobile & 5926 ms & 80.0 MB & 6.2 MB \\
        TensorFlow Lite (standard operations) & 266 ms & 12.3 MB & 1.8 MB \\
        TensorFlow Lite (manually fused custom operation) & 63 ms & 6.2 MB & 1.8 MB \\
        \stf & 128 ms & 4.2 MB & 3.6 MB\\
        \hline
    \end{tabular}
    \caption{On-device training statistics for a personalized spline model across four different implementations.}
    \label{tab:zbdperf}
\end{table*}

The results of fine-tuning this spline model to convergence are shown in Table \ref{tab:zbdperf}.
The \stf\ implementation provides the lowest memory usage, as measured by the differences in peak process memory size before and after training.
\stf\ trains to convergence faster than all but the customized TensorFlow Lite implementation.
At the time of these tests, the Swift compiler was unable to generate appropriate NEON vector instructions on Android for this model, which partially accounts for the time difference with the custom fused training operation. 

For performance reasons, it is generally impractical to use Python model code when deploying to mobile devices, so the TensorFlow Mobile and TensorFlow Lite versions of this model used Python for the server-based global model training and a binary model graph driven via Java and C++ for on-device fine tuning.
In contrast, the same Swift code defined and ran model training in both stages. This is advantageous for development and ongoing maintenance.

\section{Related Work}
\label{section:related-work}

\pt~\cite{pytorch}, Chainer~\cite{chainer}, and Autograd~\cite{autograd-thesis} have explored a define-by-run programming model in the context of the Python programming language. 
\stf\ provides a similar end-user experience, but in a statically typed programming language. 
Our AD system is orthogonal to the tensor library to allow the differentiation of arbitrary user defined types, and offers generalized support for defining derivatives alongside the original functions.

Both \tf~\cite{tensorflow} and \stf\ share portions of the underlying technology stack. 
\stf's eager execution is based directly on \tf\ Eager~\cite{tensorflow-eager}.
We rely on lazy tracing to offer opt-in support for JIT compilation to XLA. \tf\ exposes similar functionality using the \texttt{@tf.function} annotation.

JAX~\cite{jax} extends \np~\cite{numpy} with support for opt-in JIT compilation based on XLA. 
JAX traces~\cite{jaxtracing} a pure subset of \texttt{@jit} annotated Python programs into its own intermediate representation called jaxpr, and transforms these jaxprs on-the-fly or after-the-fact to perform automatic differentiation, XLA compilation, and other program transformations.

Unlike JAX's \texttt{@jit} and \tf's \texttt{@tf.function}, \stf\ supports programs that mix both tensor and arbitrary computations that are run on the host. 
This is possible with lazy tracing that seamlessly integrates into the host language without breaking the eager-like programming model.

Similar to DyNet~\cite{dynet}, we implicitly construct computation graphs on each iteration of the training or inference loops.
This allows us to support fully dynamic networks that can change architecture on each iteration, at a cost of incurring tracing overhead.
Unlike DyNet, our LazyTensor implementation automatically decides when to materialize and compile the constructed graphs. 
To minimize this overhead, we use tracing purely to support just-in-time compilation on accelerators via XLA.

The \stf\ AD system has a number of things in common with Zygote \cite{juliaAutodiff}. 
Both systems rely on source-to-source transformations to automatically generate derivatives. 
The key differentiating aspect is that \stf\ manages to do this completely statically in an AOT setting.
This makes it suitable for supporting machine learning on edge devices where the cost of tracing and JIT compilation are infeasible. 

Approaches built around the Lightweight Modular Staging framework~\cite{lms} such as Flare~\cite{flare} and Lantern~\cite{lantern} perform compilation to heterogeneous hardware through a language-integrated support for staging.
LazyTensor traces to an intermediate representation suitable for domain-specific compilers without breaking the illusion of eager execution.
In contrast, LMS requires explicit type annotations to direct the staging framework. 

\section{Conclusion}
\label{section:conclusion}

In this paper we introduced \stf, a platform for deep learning that has been validated with over 30 models.
We have demonstrated how this platform scales across a wide range of hardware envelopes, how a modern AD system can be embedded in a statically typed language, and how a lazy implementation of a \tensor\ type allows JIT-compilation via domain-specific compilers without sacrificing the eager programming model.
We have also shown how mutable value semantics addresses challenges when programming ML systems, including simple solutions to (a) reducing memory consumption on accelerators without compromising the programming model and (b) the array subscript derivative problem.
We believe there is significant promise in exploring machine learning in the context of statically-typed AOT-compiled languages.

\section*{Acknowledgements}

The authors would like to thank the many folks who made significant contributions to the Swift for TensorFlow project including Olzhas Akpambetov, Pawan Sasanka Ammanamanchi, Tanmay Bakshi, Paige Bailey, Gogul Balakrishnan, Eugene Burmako, James Bradbury, Bart Chrazczcz, Ed Connell, Mingsheng Hong, Brett Koonce, Shaddaj Laddad, Ewa Matejska, Adam Paszke, Anthony Platanios, and Alex \c{S}uhan.
We would also like to thank Jeremy Howard, Sylvain Gugger, and Rachel Thomas from fast.ai, whose collaboration substantially influenced the Swift for TensorFlow project.
A big thank you goes out to the Google Summer of Code participants (Ayush Agrawal, Victor Antony, Param Bhavsar, Karthik Ramesh Iyer, Seungjae "Ryan" Lee, and Ayushi Tiwari) and our two Google Code-In Grand Prize winners (Rick Wierenga , William Zhang).
The automatic differentiation system was heavily influenced by the work of, and conversations with Casey Chu, Conal Elliott, Roy Frostig, Matthew Johnson, Dougal Maclaurin, Bart van Merri\"{e}nboer, Gordon Plotkin, Alexey Radul, Alex Wiltschko, and Dimitrios Vytiniotis.
We also deeply appreciate the collaborations across Georgia Tech, Google, and DeepMind and would like to acknowledge the work of Mike Ando, Frank Dellaert, Dominik Grewe, Dmitri Gribenko, Tim Harley, Fan Jiang, James Keeling, Marc Lanctot, Edward Lockhart, and Julian Schrittwieser.
This work would not have been possible without support from the Swift core team and related folks including: Steve Canon, Doug Gregor, Joe Groff, Ted Kremenek, John McCall, and Slava Pestov.
Thank you to the many reviewers of this paper including Martin Abadi, James Bradbury, James Laudon, Rif A. Saurous, and Cliff Young for their invaluable feedback.
And finally, we would like to thank the amazing \stf\ community at large that has made multiple books, the majority of the models, and myriad other contributions; we couldn't have done this without you.

\bibliographystyle{mlsys2021}
\bibliography{references}

\appendix
\onecolumn

\clearpage

\section{\texttt{inout} and Mutable Value Semantics}

\begin{figure*}[h!]
\begin{multicols}{2}
\begin{minted}{swift}
func inc(_ x: inout Int) -> Bool {
  x = x + 1
  return x < 10
}

var y = 2, z = false
z = inc(&y)

print(y, z)
\end{minted}

\columnbreak

\begin{minted}{swift}
func inc(_ x0: Int) -> (Int, Bool) {
  let x = x0 + 1
  return (x, x < 10)
}

var y = 2, z = false
(y, z) = inc(y)

print(y, z)
\end{minted}

\end{multicols}
\caption{Rewriting pass-by-\texttt{inout} as pass-by-value plus assignment. Both programs print ``3 true.'' A call using
  \texttt{inout} (left column) can be transformed into an equivalent call using pass-by-value (right column), by returning an updated value for each \texttt{inout} parameter to the original function and assigning into corresponding \texttt{inout} argument(s) at the call site. This equivalence demonstrates that  \texttt{inout} does not introduce reference semantics.}
\label{fig:inout}
\end{figure*}

\section{Array Subscript Differentiation}
\label{app:arr-ad}

\begin{figure*}
\begin{scriptsize}
\begin{minted}{swift}
// Example operation to differentiate.
func myOp(values: [Float], a: Int, b: Int) -> Float {
  return values[a] + values[b]
}

// Functional representation

// subscript read with explicit pullback, functional style.
func subscriptWithFunctionalPullback(values: [Float], index: Int) -> (
  value: Float,
  pullback: (Float) -> [Float]  // Functional formulation.
) {
  let size = values.count  // Optimization: don't capture whole array, just size.
  return (values[index], { dx in
    var tmp = Array(repeating: Float(0), count: size)  // Allocates O(n) memory!
    tmp[index] = dx
    return tmp
  })
}

// Returns a new array where each element is the sum of the corresponding
// elements of `a` and `b`.
func sumArraysHelper(_ a: [Float], _ b: [Float]) -> [Float] {
  precondition(a.count == b.count)
  var result = Array(repeating: Float(0), count: a.count)
  for i in 0..<a.count { result[i] = a[i] + b[i] }
  return result
}

// The operation and its corresponding pullback written in functional style.
func myOpWithFunctionalPullback(values: [Float], a: Int, b: Int) -> (
  value: Float, pullback: (Float) -> [Float]
) {
  let (aVal, aPb) = subscriptWithFunctionalPullback(values, a)
  let (bVal, bPb) = subscriptWithFunctionalPullback(values, b)
  let result = aVal + bVal
  return (result, { dx in
    let dA = aPb(dx)  // O(n), allocates O(n) memory.
    let dB = bPb(dx)  // O(n), allocates O(n) memory.
    return sumArraysHelper(dA, dB)  // O(n)
  })
}

// Value semantic representation

// subscript read with explicit pullback, value semantic style
func subscriptWithMutablePullback(values: [Float], index: Int) -> (
  value: Float,
  pullback: (Float, inout [Float]) -> Void  // Value semantic formulation.
) {
  return (values[index], { dx, dValues in
    dValues[index] += dx  // Constant time!
  })
}

// The operation and its corresponding pullback written value semantic style.
func myOpWithMutablePullback(values: [Float], a: Int, b: Int) -> (
  value: Float, pullback: (Float, inout [Float]) -> Void
) {
  let (aVal, aPb) = subscriptWithMutablePullback(values, a)
  let (bVal, bPb) = subscriptWithMutablePullback(values, b)
  return (aVal + bVal, { dx, dValues in
    aPb(dx, &dValues)  // Constant time.
    bPb(dx, &dValues)  // Constant time.
  })
}

\end{minted}
\end{scriptsize}
\caption{A simple, differentiable function using array subscripts, and two alternative AD formulations explicitly written, one in a pure-functional paradigm, and a second leveraging mutable value semantics which reduces derivative complexity from $O(n)$ to $O(1)$.}
\label{fig:arr-ad}
\end{figure*}

Figure \ref{fig:arr-ad} contains a reference program, and two explicitly-written implementations of JVPs, one in the immutable-functional style, and one leveraging mutable value semantics.
The operation to differentiate is $O(1)$.
The functional pullback formulation runs in $O(n)$ time, where $n$ is the size of \texttt{values}.
The value semantic formulation runs in constant time, irrespective of the size of \texttt{values}.
For an executable form of this appendix, please see: \url{https://colab.research.google.com/github/tensorflow/swift/blob/master/notebooks/value_semantics/02_value_semantics_and_autodiff.ipynb} that walks through the functional AD formulation before introducing this case study.

\end{document}